\documentclass[runningheads]{llncs}

\usepackage{graphicx}%
\usepackage{multirow}%
\usepackage{amsmath,amssymb,amsfonts}%
\usepackage{mathrsfs}%
\usepackage[title]{appendix}%
\usepackage{xcolor}%
\usepackage{textcomp}%
\usepackage{manyfoot}%
\usepackage{booktabs}%
\usepackage{algorithm}%
\usepackage{algorithmicx}%
\usepackage{algpseudocode}%
\usepackage{listings}%
\usepackage{float}
\usepackage{balance}
\usepackage{enumitem}
\usepackage{tabularx}
\usepackage{ragged2e}
\usepackage{booktabs}
\setlist{nosep}
\usepackage{wrapfig}

\usepackage{pifont}

\usepackage{subcaption}
\usepackage{balance}
\usepackage{colortbl}
\definecolor{lightgray}{gray}{0.9}

\usepackage[accsupp]{axessibility}
\usepackage[pagebackref,breaklinks,colorlinks,citecolor=blue]{hyperref}
\usepackage{orcidlink}

\newcolumntype{L}[1]{>{\raggedright\arraybackslash}p{#1}}
\newcolumntype{R}[1]{>{\raggedright\arraybackslash}p{#1}}

\newcommand{\repthanks}[1]{\textsuperscript{\ref{#1}}}
\makeatletter
\patchcmd{\maketitle}
{\def\thanks}
{\let\repthanks\repthanksunskip\def\thanks}
{}{}
\patchcmd{\@maketitle}
{\def\thanks}
{\let\repthanks\@gobble\def\thanks}
{}{}
\newcommand\repthanksunskip[1]{\unskip{}}
\makeatother

\makeatletter
\DeclareRobustCommand\onedot{\futurelet\@let@token\@onedot}
\def\@onedot{\ifx\@let@token.\else.\null\fi\xspace}

\begin{document}

\title{Toward 360-Degree Indoor Panorama Editing via Tuning-Free Diffusion Model with Refocusing Cross-Attention}

\titlerunning{FocusDiff: Toward 360-Degree Indoor Panorama Editing}

\author{
Dinh-Khoi Vo\inst{1,2}\orcidlink{0000-0001-8831-8846} \and
Nhut-Thanh Le-Hinh\inst{1,2}\orcidlink{0009-0001-7199-3180} \and
Viet-Tham Huynh\inst{1,2}\orcidlink{0000-0002-8537-1331} \and
Tam V. Nguyen\inst{3}\orcidlink{0000-0003-0236-7992} \and
Minh-Triet Tran\inst{1,2}\orcidlink{0000-0003-3046-3041} \and
Trung-Nghia Le\thanks{Corresponding author.}\inst{1,2}\orcidlink{0000-0002-7363-2610}}

\institute{University of Science, Ho Chi Minh, Vietnam \and
Vietnam National University, Ho Chi Minh, Vietnam \and University of Dayton, Ohio, United States \\
\email{\{vdkhoi,lhnthanh,hvtham\}@selab.hcmus.edu.vn}\\
\email{tamnguyen@udayton.edu}\\
\email{\{tmtriet,ltnghia\}@fit.hcmus.edu.vn}}

\authorrunning{Dinh-Khoi Vo et al.}

\maketitle

\begin{abstract}
Zero-shot text-guided diffusion has significantly advanced image editing; however, its practical usability remains constrained by three persistent challenges: prompt brittleness that requires meticulous prompt engineering, spillover edits that unintentionally affect non-target regions, and failures on small or cluttered objects caused by limited fine-grained supervision in training data. We propose \textbf{FocusDiff} (\emph{Target-Aware Refocusing for Tuning-Free Diffusion Editing}), a tuning-free framework for precise and region-specific image manipulation based on refocusing cross-attention. Given a target region obtained through automated segmentation or manual selection, FocusDiff applies selective blurring to non-edit areas to guide attention toward the masked region while accurately transferring the object's identity, structure, and appearance to the edited output. Integrated context-preserving modules further ensure background fidelity and global coherence, enabling accurate edits from simple text prompts in a single pass. We also extend FocusDiff to 360-degree indoor panorama editing and demonstrate its effectiveness within virtual reality environments. Extensive experiments on our localized editing benchmark LIMB, comprising 30 multi-object images and 100 annotated examples including challenging small-object cases, show that FocusDiff outperforms existing zero-shot editors in text–image alignment and background preservation, achieving superior precision, photorealism, and usability. The project page is available at \url{https://vdkhoi20.github.io/FocusDiff}.


\end{abstract}

\begin{figure}
    \centering
    \includegraphics[width=\textwidth]{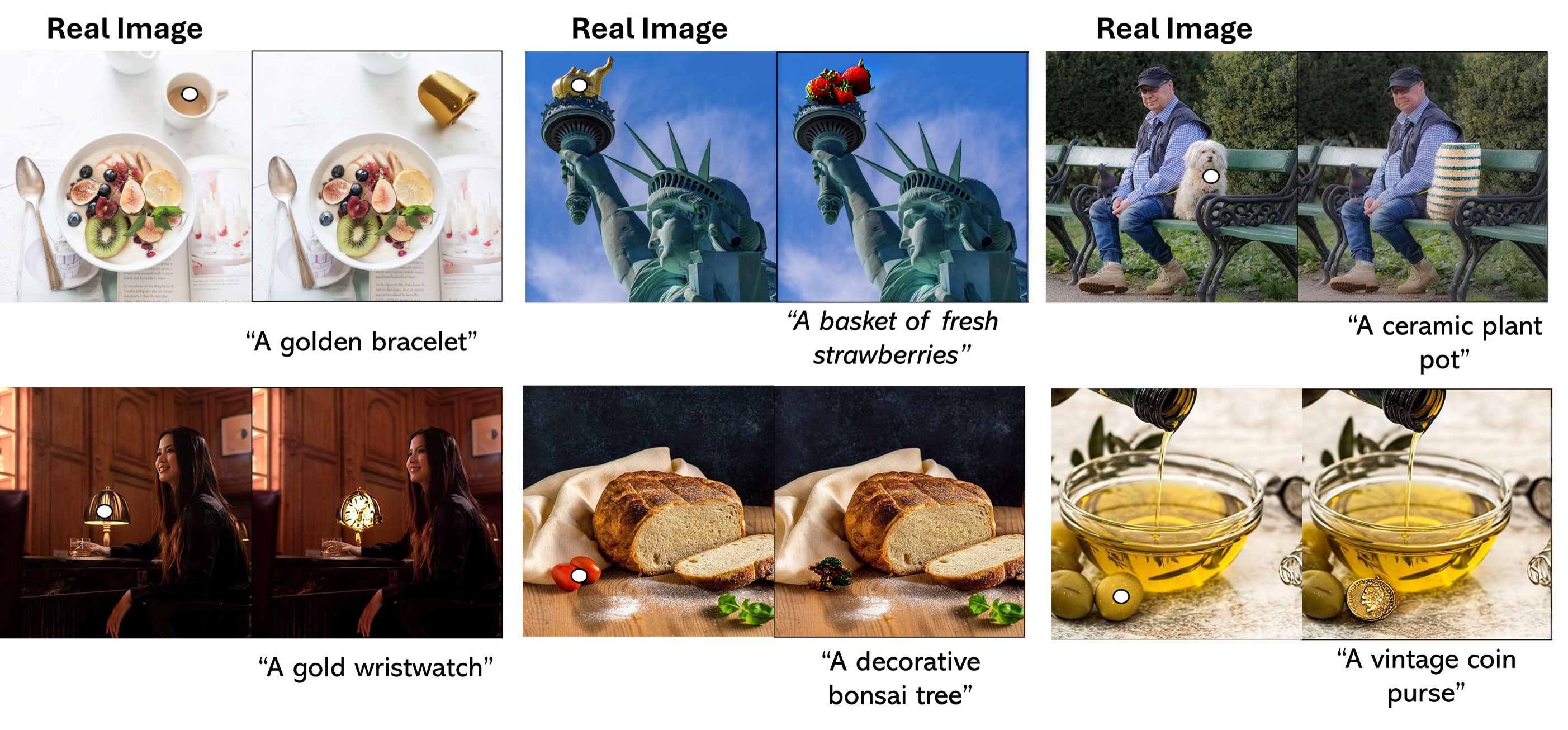}
    \vspace{-5mm}
    \caption{FocusDiff enables precise, region-specific edits from simple prompts without fine-tuning, accurately modifying small objects while preserving surrounding content and visual coherence. The examples illustrate its effectiveness on challenging scenes with small targets and complex backgrounds.}
    \label{fig:teaser}
    \vspace{-5mm}
\end{figure}

\section{Introduction}
Recent progress in text-to-image (T2I) generation has revolutionized AI-assisted content creation. Large-scale diffusion models~\cite{rombach2022highresolution} have shown remarkable success in generating high-quality, diverse images conditioned on textual descriptions~\cite{dhariwal2021diffusion}. These advancements have facilitated significant improvements in text-driven image editing~\cite{hertz2022prompt,Tumanyan_2023_CVPR,parmar2023zeroshot}, enabling users to modify visual content using natural language instructions.  

Despite these advancements, real-world image editing remains a challenging problem~\cite{vdkhoi_chi2024,parmar2023zeroshot}. The key objective is to modify specific objects or regions within an image while maintaining structural coherence and preserving background consistency. However, existing approaches often struggle with localized editing as illustrated in Fig.~\ref{fig:teaser}, as they tend to focus on the most visually salient objects rather than those explicitly intended by the user. This issue arises from the training paradigm of diffusion models~\cite{rombach2022highresolution}, which rely on large-scale image–caption datasets~\cite{schuhmann2021laion} that typically lack fine-grained object descriptions and representations of multi-object interactions. Furthermore, real-world scenes often include numerous or small objects with complex spatial relationships, making precise manipulation particularly difficult. In addition, fully fine-tuning large-scale generative models~\cite{ruiz2023dreambooth,gal2022textual,brooks2022instructpix2pix} is computationally expensive, which limits their practicality for research and real-world deployment.  

Since training diffusion models on large-scale datasets is computationally expensive, recent work has shifted toward tuning-free strategies that reuse pre-trained text-to-image (T2I) models such as Stable Diffusion~\cite{rombach2022highresolution}. This line of research, commonly referred to as \textit{zero-shot image editing}, exploits frozen model weights to perform edits without fine-tuning~\cite{meng2022sdedit,brack2024ledits++,cao_2023_masactrl,liu2024towards,li2024zone,Tumanyan_2023_CVPR}. While these methods enable efficient and scalable image manipulation, they remain limited in handling localized edits, where changes must be accurately confined to a specific region without disturbing surrounding content. Several methods~\cite{hertz2022prompt,avrahami2023blended,couairon2022diffedit} attempt to address localized editing by blending the original latent noise with the edited noise, or by directly blending on the image. However, these approaches overlook the interactions between foreground and background and struggle with small objects~\cite{vo2025cpam,mao2024mag,Simsar_2025_WACV,li2024zone}, often resulting in rigid, less natural edits or, in some cases, no visible changes at all.

To overcome the limitations of existing diffusion-based editing methods, we propose \textbf{Target-Aware Refocusing for Tuning-Free Diffusion Editing (FocusDiff)}, a novel framework that enables precise, region-specific modifications in complex images containing multiple objects. These localized edits are guided by a simple text prompt without requiring fine-tuning or optimization. Inspired by the joint refinement of cross- and self-attention~\cite{vo2025cpam,cao_2023_masactrl}, FocusDiff retains background fidelity while directing the model’s focus toward the target region. Specifically, we first identify the region of interest via automated segmentation or manual selection, and then apply \textit{refocusing cross-attention}, where non-editing areas are blurred to emphasize the model’s attention to the target region while consistently trasferring the object’s identity, structure, and appearance to the edited image. At the same time, we regulate self-attention and cross-attention by reusing the context-preserving modules from CPAM~\cite{vo2025cpam}, which stabilize background regions, prevent unintended distortions and maintain the visual coherence. 

To evaluate localized editing, we introduce the Localized Image Manipulation Benchmark (LIMB), a subset of PIE-Bench~\cite{ju2023direct} containing 30 multi-object images with 100 annotated examples and prompts, providing a standardized testbed for assessing fine-grained, region-specific image edits. Experimental results demonstrate that FocusDiff surpasses existing zero-shot SOTA methods in localized editing by offering a more efficient, intuitive, and user-friendly approach. 

Beyond conventional 2D image editing, we further extend FocusDiff to panoramic imagery, demonstrating its capability to handle challenging scenarios involving multiple small objects whose sizes are negligible relative to the overall image. This extension also highlights FocusDiff’s potential for applications in Virtual Reality (VR) and immersive media environments.

Key contributions of our work are as follows:  
\begin{itemize}
    \item We introduce a novel zero-shot localized editing framework that enables precise, region-specific modifications in complex images containing multiple objects. The proposed FocusDiff performs edits guided by simple textual instructions in a single pass while preserving the integrity of non-edited regions, without requiring model fine-tuning or optimization. 

    \item We proposed a refocusing cross-attention mechanism that sharpens editing precision by steering attention to the target region via selective blurring of non-editing areas, while transferring the object’s identity, structure, and appearance to the edited result.

    \item FocusDiff integrates context-preserving strategies from existing image editing methods, ensuring background preservation while maintaining flexible modification capabilities.

    \item We extend our framework to panoramic 360-degree indoor images for Virtual Reality (VR) environments, demonstrating its applicability to immersive scenarios beyond conventional 2D imagery. 

    \item We present LIMB dataset with 30 multi-object images and 100 annotated examples, serving as a benchmark for fine-grained, region-specific image edits.
    
    \item Extensive experiments show that FocusDiff outperforms zero-shot SOTA approaches in localized image editing, achieving superior accuracy, usability, and photorealism.  
\end{itemize}

\section{Related Work}
\label{sec:related}

\subsection{Zero-shot image editing}

Zero-shot image editing modifies images during denoising without fine-tuning by leveraging pre-trained T2I models with frozen weights~\cite{rombach2022highresolution,meng2022sdedit,brack2024ledits++}. Most methods rely on attention mechanisms to preserve image structure~\cite{hertz2022prompt,cao_2023_masactrl,liu2024towards,parmar2023zeroshot,Tumanyan_2023_CVPR}, using self-attention swaps~\cite{Tumanyan_2023_CVPR,liu2024towards} or cross-attention replacements~\cite{hertz2022prompt}. However, these approaches struggle with significant object variations or require semantic similarity between prompts. Alternative methods introduce noise-based guidance~\cite{meng2022sdedit}, fast inversion techniques~\cite{brack2024ledits++}, or large-scale text banks~\cite{parmar2023zeroshot}. MasaCtrl~\cite{cao_2023_masactrl} improves attention retention but lacks flexibility in controlling background-foreground interactions. Meanwhile, CPAM~\cite{vo2025cpam} integrates self- and cross-attention to achieve precise, localized editing of non-rigid objects while maintaining background integrity.


Existing zero-shot localized editing methods, such as Blended Diffusion~\cite{avrahami2023blended} enables localized editing through gradient guidance based on CLIP loss but often overlooks foreground–background interactions, resulting in rigid and inconsistent outputs. Other methods perform local edits by blending latent noise or directly compositing image regions~\cite{hertz2022prompt,avrahami2023blended,couairon2022diffedit,li2024zone}, while recent strategies enhance localization by guiding attention toward target regions and preserving background consistency~\cite{vo2025cpam}. Despite these advances, existing zero-shot approaches still struggle to reliably modify small objects or fine details, frequently producing incomplete edits or unintended changes in surrounding areas. Our method addresses this challenge by refocusing the model’s attention on the target region through a lightweight blurring mechanism, enabling precise, region-specific edits while preserving the integrity of the background.

\subsection{Panoramic editing in immersive environment}
Recent research has devoted considerable effort to integrating panoramic images into Virtual Reality (VR) environments. Most existing work focuses on enhancing visual realism and user comfort~\cite{LayerPano3D,Real_Time_Omnidirectional,PanoTrace,Huang_2025_CVPR}, or on embedding panoramic images within real-world contexts, such as digital diaries in VR~\cite{Digital_dỉaries}. In contrast, relatively little attention has been given to interactive editing of panoramic images within VR, a challenge frequently highlighted in the literature~\cite{tukur2025panoramic}. Tsai et al.~\cite{DreamCraft} explored VR scene reconstruction by editing panoramic images through object removal, modification, and style transfer. However, the resulting performance remains limited and does not provide a broadly practical solution for interactive VR editing.

\section{Proposed Method}

\subsection{Overview of FocusDiff} 

\begin{figure}[!t]
    \centering
    \includegraphics[width=\textwidth]{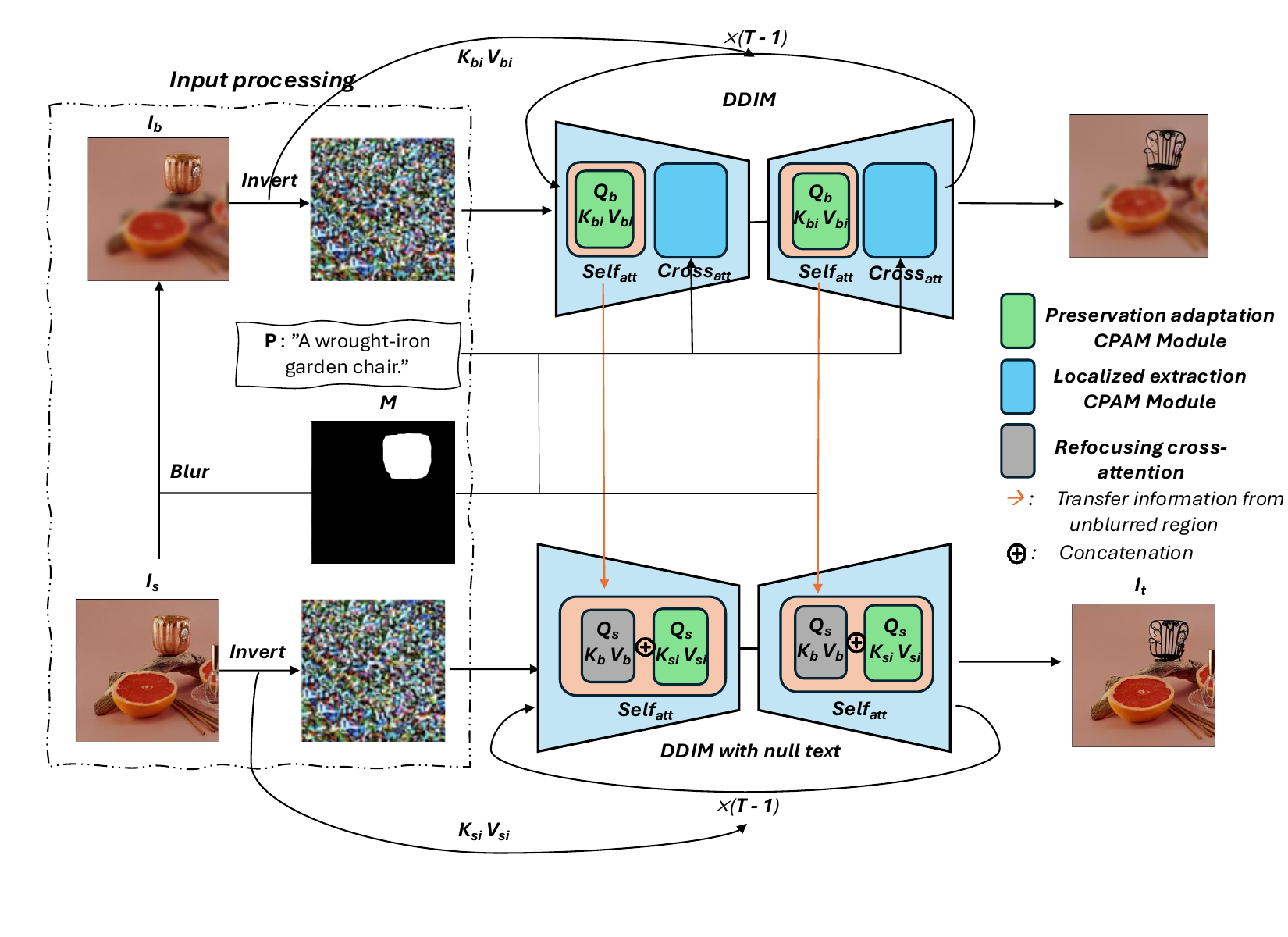}
    \vspace{-12mm}
    \caption{Pipeline of our proposed Target-Aware Refocusing for Tuning-Free Diffusion Editing (FocusDiff). Given a source mask $M$, the input image is blurred and processed in two parallel flows. During denoising, the proposed Refocusing Cross-Attention module transfers rich semantic cues and fine structural details of edited object from the blurred latent to the edited latent, sharpening the target region, while CPAM's modules~\cite{vo2025cpam} are utilized to preserve background harmony, yielding precise, natural, and context-aware localized edits.}
    \label{fig:pipeline}
    \vspace{-5mm}
\end{figure}

Figure~\ref{fig:pipeline} illustrates the overall pipeline of our proposed \textit{Target-Aware Refocusing for Tuning-Free Diffusion Editing (FocusDiff)}, a framework for precise region-based image manipulation. Given an input image $I_s$ and a source object mask $M$, obtained through various strategies such as manual drawing, click-based extraction, or text-prompt guidance via SAM~\cite{kirillov2023segment}, our method generates a modified image $I_t$ that aligns with the semantics of a target text prompt $P$ while preserving the structure and content of non-edited regions. Specifically, FocusDiff consists of three components: (i) {Refocusing Cross-Attention}, which sharpens editing precision by blurring non-target areas while transferring the object’s identity, structure, and appearance; (ii) Context-Preserving Integration, ensuring background fidelity and structural consistency; and (iii) a Panoramic Extension, enabling 360-degree indoor panorama editing for Virtual Reality (VR).

\subsection{Refocusing cross-attention}   
To strictly enforce the model’s attention on the target region within the cross-attention module, we introduce an efficient strategy that suppresses non-editing areas through lightweight blurring, thereby guiding focus toward the masked region while transferring the object’s identity, structure, and appearance back to the edited image during the denoising process.

Specifically, we first invert the original image $I_s$ and the blurred image $I_b$ into the latent noise space using DDIM inversion~\cite{song2020denoising}, which reconstructs the noise deterministically. We then feed these latents into the denoising process, at each timestep $t$ the object’s identity, structure, and appearance are extracted and then transferred back to the edited latent.

At each denoising step $t$, we denote the query–key–value feature triplets from the blurred and source latent noises as $(Q_b, K_b, V_b)$ and $(Q_s, K_s, V_s)$ within the U-Net’s self-attention, respectively. The semantic representation of the target object, $SC_{ob}$, is extracted through mask-guided attention:

\begin{equation}
\begin{aligned}
SC_{ob}^{blurred} = \text{Att}(Q_b, K_b, V_b; M), \\
SC_{ob}^{edited} = \text{Att}(Q_s, K_b, V_b; M),
\end{aligned}    
\end{equation}

where $\text{Att}(\cdot)$ denotes the attention operation and $M$ restricts computation to the masked region and $SC$ denotes as semantic content. This design allows us not only to perform localized editing within the masked region but also to transfer the object’s semantic identity and structural details from the blurred latent noise to the edited latent noise, thereby achieving precise, non-rigid modifications while preserving consistency and harmony with the surrounding background.

\begin{figure}[!t]
    \centering
    \includegraphics[width=\linewidth]{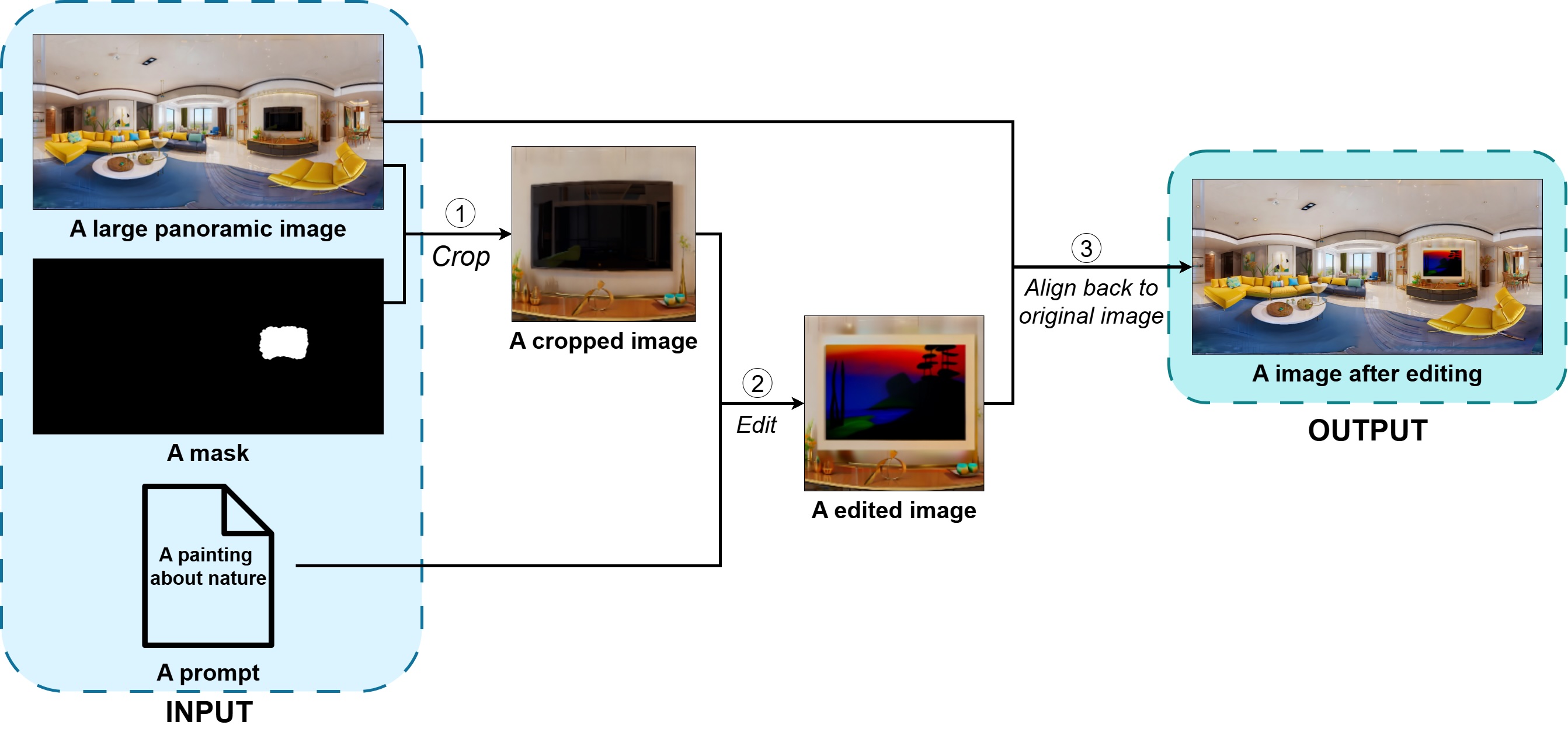}
 
    \caption{Overview of our panoramic image editing pipeline. Given a large panoramic image, our system first obtains a mask either through user selection (e.g., manual drawing or click-based tools) or automatic segmentation using a vision-language model such as SAM~\cite{kirillov2023segment}. We then crop a region of interest around the masked area, apply the editing process to this localized region for precise modification, and finally align the edited region back into the original panoramic image to preserve global consistency and avoid artifacts.}

    \label{fig:panoramic}
    \vspace{-5mm}
\end{figure}

\subsection{Integration of context-preserving}  
Preserving background regions in both blurred and edited latent noises is crucial, as information gradually degrades during the denoising process. To mitigate this, we incorporate a preservation adaptation module and a localized extraction module~\cite{vo2025cpam}, which maintain background fidelity and structural consistency across timesteps in both latent flows.

At each timestep $t$, we extract two sets of key–value feature pairs from the intermediate latent noises: $(K_{si}, V_{si})$ obtained from the inversion of $I_s$, and $(K_{bi}, V_{bi})$ obtained from the inversion of $I_b$. These features are then reintegrated into the denoising process to guide reconstruction, thereby preserving semantic identity and fine structural details. The formulation of preservation adaptation modules as follow:
\begin{equation}
\begin{aligned}
SC_{bg}^{blurred} &= \text{Att}(Q_b, K_{bi}, V_{bi}; 1-M), \\
SC_{bg}^{edited} &= \text{Att}(Q_s, K_{si}, V_{si}; 1-M), \\
SC^{blurred} &= M \cdot SC_{ob}^{blurred} + (1-M) \cdot SC_{bg}^{blurred}, \\
SC^{edited} &= M \cdot SC_{ob}^{edited} + (1-M) \cdot SC_{bg}^{edited}.
\end{aligned}
\end{equation}

Finally, to ensure that the text prompt $P_t$ only drives modifications within the masked region, we integrate a localized extraction module~\cite{vo2025cpam}. This module conditions the masked area with $P_t$, while the unmasked background is conditioned on a null prompt, thereby preventing undesired global alterations.

\begin{figure}[!t]
    \centering
    \includegraphics[width=0.7\linewidth]{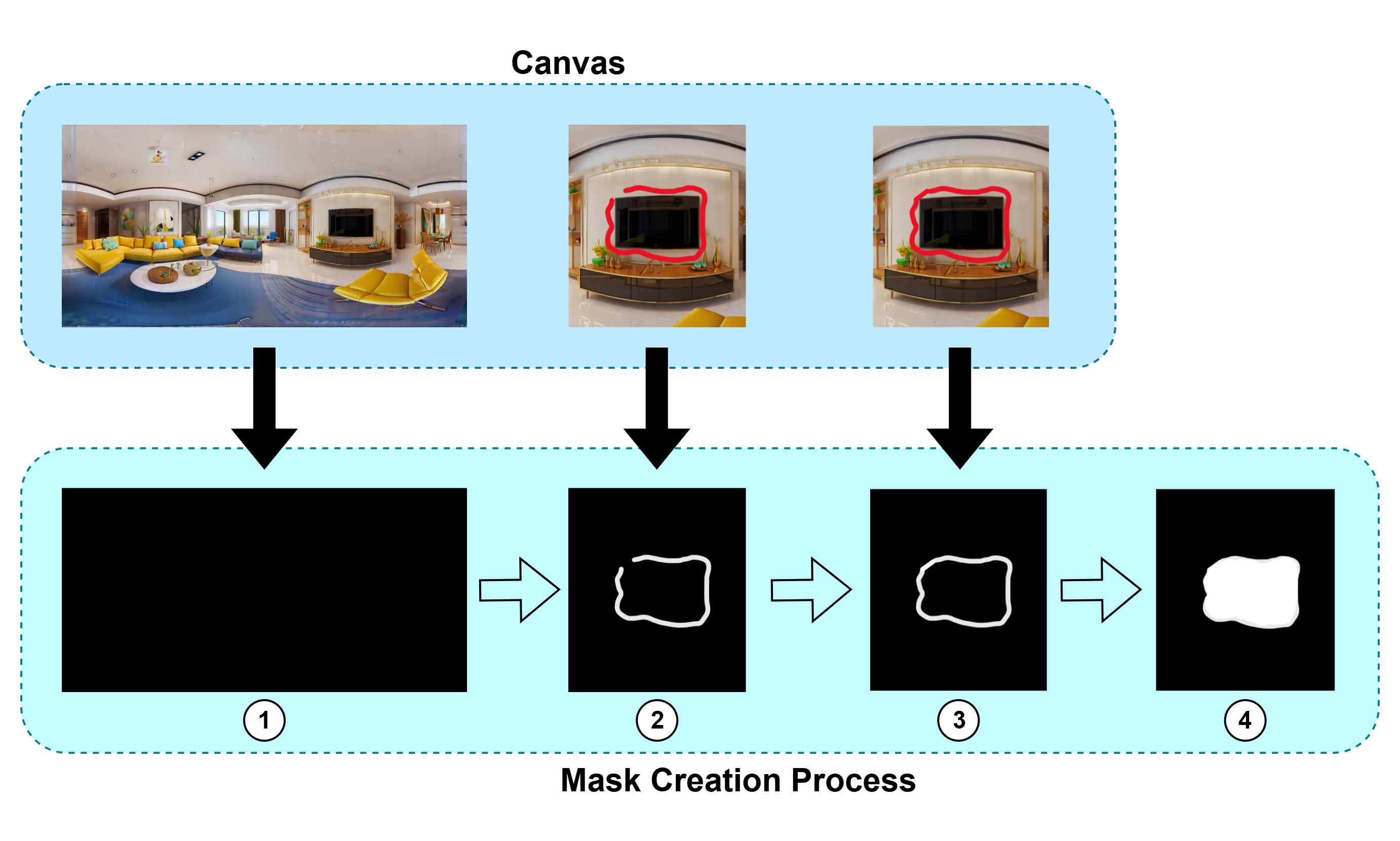}
    \vspace{-5mm}
    \caption{Mask generation workflow. (1) A black mask is initialized with the same resolution as the source panorama. (2) As the user sketches the target object’s contour, the drawn pixels are marked in white. (3) An auto-connect algorithm links the first and last stroke points to form a closed contour. (4) A flood-fill operation propagates white pixels within the enclosed region, producing the final binary mask.}
    \label{fig:mask_generation_workflow}
\end{figure}

\begin{figure}[!t]
    \centering
    \begin{subfigure}{0.47\linewidth}
        \centering
        \includegraphics[width=\linewidth]{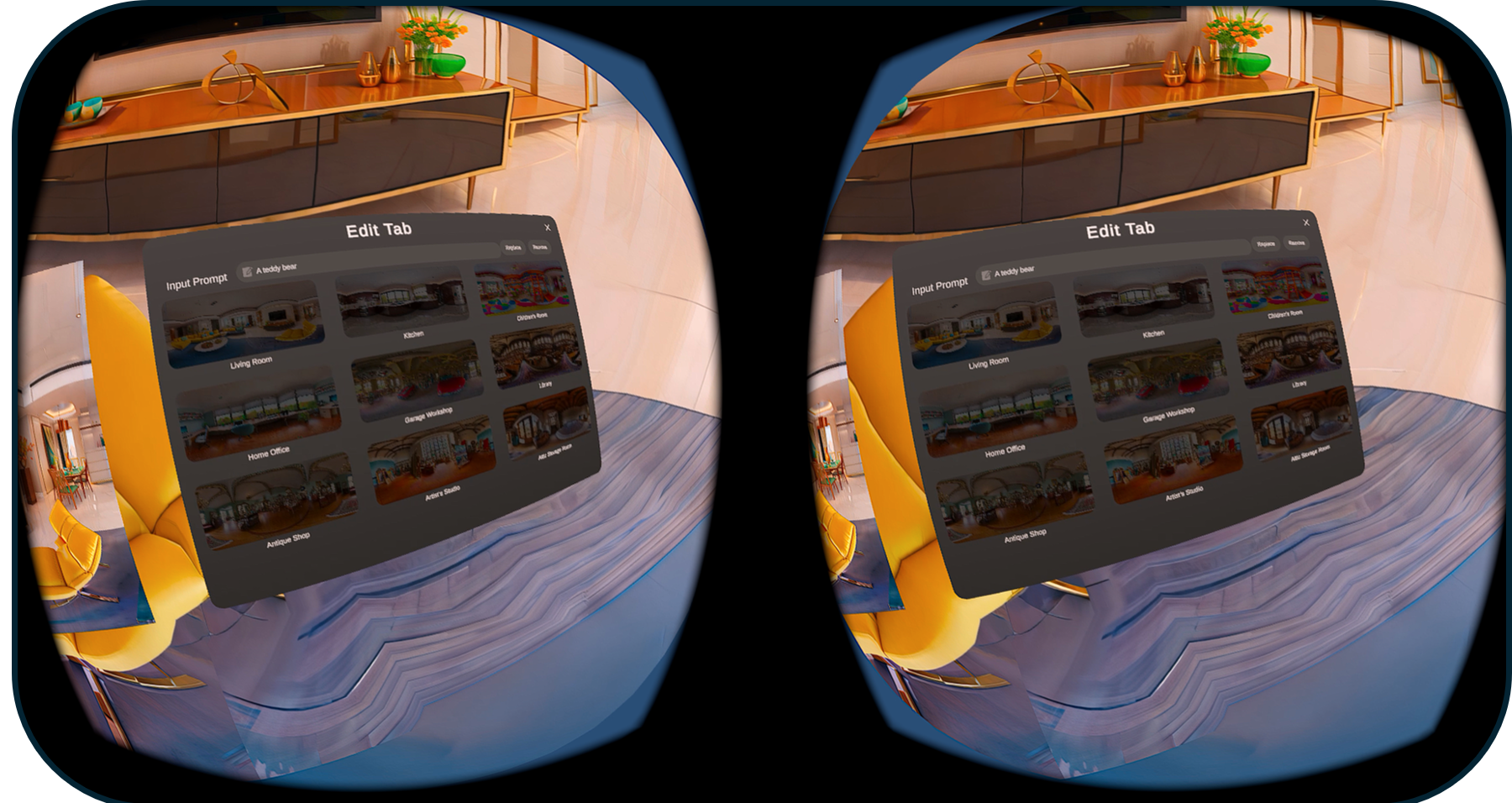}
        \caption{\textit{Editing Panel}: displays the list of available panoramic images, a prompt text field, and action buttons for object replacement or removal.}
        \label{fig:VR_UI_B}
    \end{subfigure}
    \hfill
    \begin{subfigure}{0.47\linewidth}
        \centering
        \includegraphics[width=\linewidth]{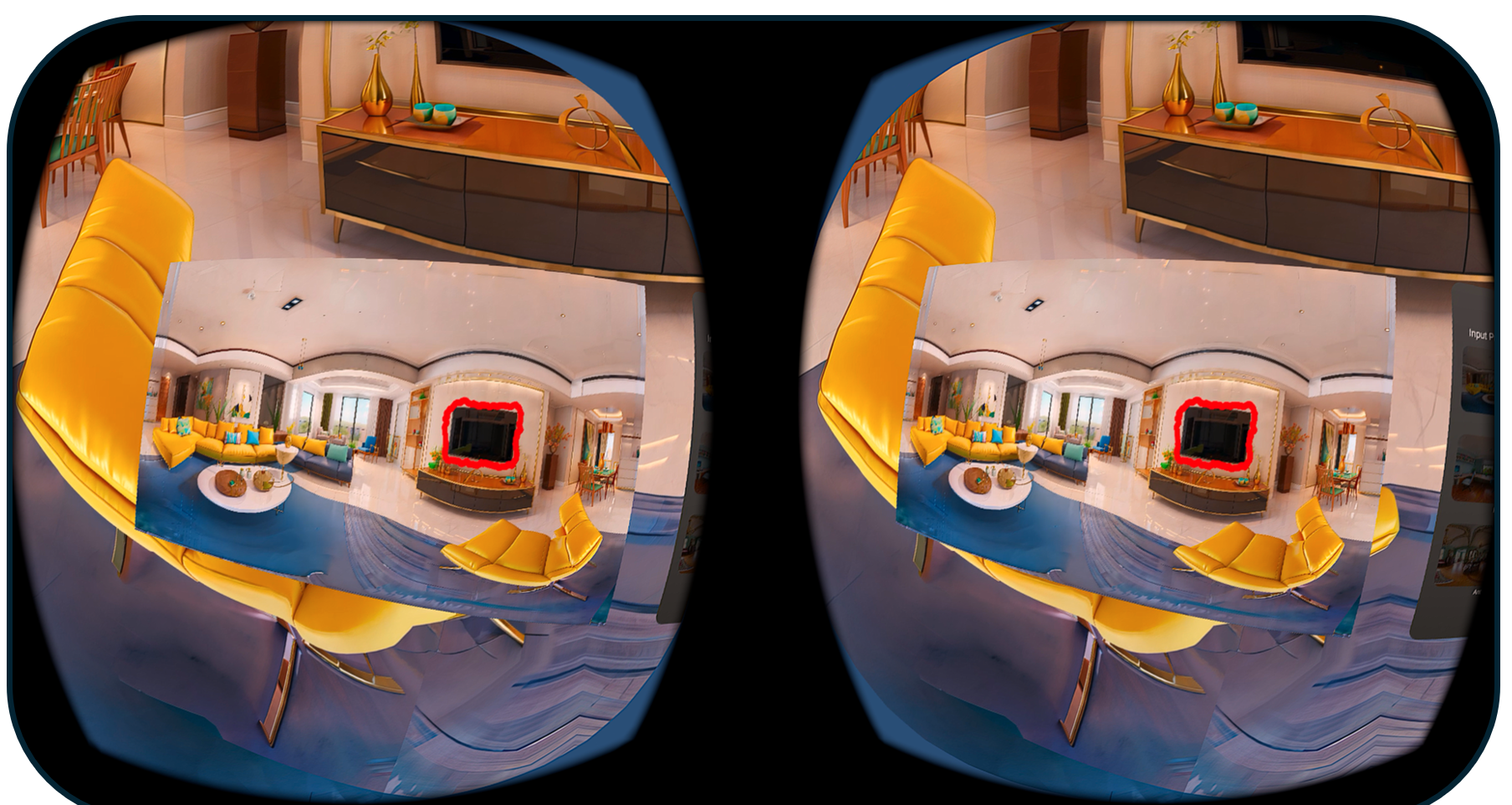}
        \caption{\textit{Mask Generation Canvas}: enables users to draw directly on the scene to define regions for targeted edits.\\}
        \label{fig:VR_UI_C}
    \end{subfigure}
    \caption{Interface of the VR-based panoramic editing system, illustrating key components: 
    Editing Panel, and Mask Generation Canvas.}
    \label{fig:VR_tool_UI}
    \vspace{-5mm}
\end{figure}

\subsection{VR-based panoramic image editing}

Editing small, localized regions within large panoramic images is challenging due to the scale difference between the entire panorama and the targeted area. To address this, we develop a \textbf{VR-based panoramic editing} system, built upon our FocusDiff framework, which integrates a localized cropping strategy with an interactive virtual reality interface.

\textbf{Processing pipeline. }
As illustrated in Fig.~\ref{fig:panoramic}, our panoramic editing pipeline begins by generating a mask for the target region, either through user interaction (e.g., manual drawing or click-based tools) or automatic segmentation using a vision-language model such as SAM~\cite{kirillov2023segment}. A region of interest surrounding the masked area is then cropped to retain sufficient contextual information while reducing computational overhead. Edits are applied to this localized region using our FocusDiff framework, enabling precise and context-aware modifications. Finally, the edited region is seamlessly aligned and blended back into the original panorama, ensuring global consistency, structural integrity, and a natural visual appearance.

As illustrated in Fig.~\ref{fig:mask_generation_workflow}, the mask generation process begins by initializing a black mask with the same resolution as the source panorama. When the user sketches the contour of the target object, the corresponding pixels are marked as white. An auto-connect algorithm then links the initial and final stroke points to form a closed contour. Finally, a flood-fill algorithm~\cite{BURTSEV1993549} propagates the white pixels within the enclosed boundary, resulting in a complete binary mask.

\textbf{Interactive VR interface. }
To provide an intuitive and immersive editing experience, we develop a user-friendly VR interface for panoramic image interaction (Fig.~\ref{fig:VR_tool_UI}). The system supports the following key functionalities:

\begin{itemize}
\item \textbf{Spherical Rendering:} The input panorama is projected onto a virtual sphere, allowing users to explore and edit the scene in an immersive 360-degree environment.
\item \textbf{Mask Generation:} A brush tool enables users to draw masks directly on the panoramic canvas for precise and localized editing.
\item \textbf{Object Replacement:} A text prompt field allows users to describe the object to be replaced. The system then replaces the masked object according to the user’s specification.
\item \textbf{Object Removal:} Users can remove unwanted objects within the masked area, with the surrounding context automatically inpainted to ensure visual coherence.
\item \textbf{Edit Preview:} The modified panorama is re-rendered in real time on the virtual sphere, allowing users to preview edits instantly and iteratively refine results.
\end{itemize}

\section{Experiments}


\subsection{Localized image manipulation benchmark (LIMB)} 
\label{imba}

To evaluate localized editing capabilities, we introduce the Localized Image Manipulation Benchmark (LIMB), a subset of PIE-Bench~\cite{ju2023direct} curated for more complex scene edits. LIMB consists of 30 images selected from PIE-Bench, specifically chosen for their multiple-object compositions. Each image is annotated with a mask and corresponding prompts to guide localized modifications. In total, LIMB provides 100 examples, covering a diverse range of editing tasks. This benchmark serves as a standardized testbed for assessing fine-grained, region-specific manipulations in image editing models.

\subsection{Quantitative and qualitative evaluation}

\begin{table}[!t]
\centering
\caption{Comparison with state-of-the-art methods on the LIMB benchmark using CLIPScore to measure text-image alignment (higher is better) and LPIPS to evaluate background preservation (lower is better). The two bottom rows demonstrates the generalization capability of FocusDiff across advanced diffusion backbones.} 
\label{tab:quantitative}
\begin{tabular}{lcc}
\toprule
\textbf{Method} & \textbf{CLIPScore} $\uparrow$ & \textbf{LPIPS} $\downarrow$ \\
\midrule
MasaCtrl~\cite{cao_2023_masactrl} & 20.12  & 0.280  \\
Blended-Diffusion~\cite{avrahami2023blended} & 27.43  & 0.156  \\
DiffEdit~\cite{couairon2022diffedit} & 27.75  & 0.148  \\
LEDITS++~\cite{brack2024ledits++} & 32.76  & 0.103  \\
CPAM~\cite{vo2025cpam} & 33.45  & 0.101  \\
\textbf{FocusDiff-SD1.5 (Ours)} & 35.85  & 0.099  \\
\textbf{FocusDiff-SD2.1 (Ours)} & 35.61  & 0.068  \\
\textbf{FocusDiff-SDXL (Ours)} & \textbf{36.48}  & \textbf{0.064}  \\
\bottomrule
\end{tabular}
\vspace{-3mm}
\end{table}

\begin{figure}[!t]
    \centering
    \includegraphics[trim={0 7.3cm 0 0},clip, width=\textwidth]{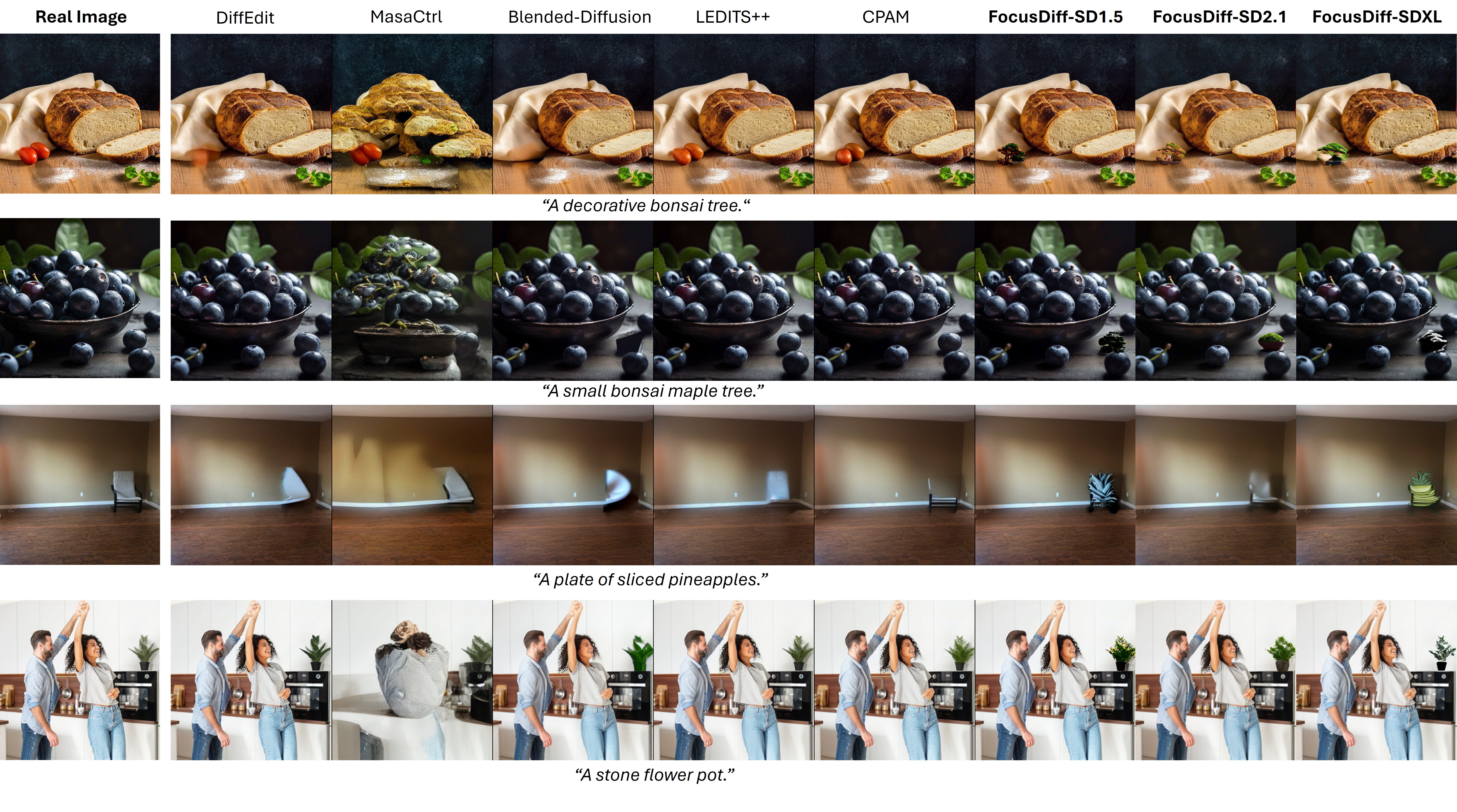}
    \vspace{-5mm}
    \caption{Qualitative comparison of zero-shot localized editing methods. Competing approaches often fail to accurately modify small or specific objects or introduce unintended background changes. In contrast, FocusDiff produces fine-grained, region-specific edits with faithful background preservation and seamless visual realism cross different backbones.}
    \label{fig:qualitative}
    \vspace{-5mm}
\end{figure}

We evaluated the proposed FocusDiff framework on the LIMB benchmark (Sec.~\ref{imba}) and compared it against leading zero-shot localized image editing methods, including MasaCtrl~\cite{cao_2023_masactrl}, Blended-Diffusion~\cite{avrahami2023blended}, DiffEdit~\cite{couairon2022diffedit}, LEDITS++~\cite{brack2024ledits++}, and CPAM~\cite{vo2025cpam}, utilizing identical manually annotated masks and a baseline configuration of 50 denoising steps with a guidance scale of 10 on Stable Diffusion v1.5. To fully demonstrate the architectural scalability and generality of our tuning-free framework, we also extended FocusDiff to more advanced backbones, including Stable Diffusion v2.1 and SDXL, without modifying their frozen internal structures. As demonstrated in the two bottom rows of Tab.~\ref{tab:quantitative}, our framework adapts seamlessly to these larger architectures and benefits substantially from their enhanced model capacity. Specifically, FocusDiff-SD2.1 significantly reduces the background LPIPS error to 0.068 due to its refined language representations, while FocusDiff-SDXL achieves the top overall performance, delivering an outstanding LPIPS of 0.064 and pushing text-image alignment to a CLIPScore of 36.48. This prominent advancement stems from the higher density of attention layers in SDXL, which provides richer semantic interactions that amplify the efficacy of our refocusing cross-attention mechanism, thereby confirming that FocusDiff possesses a robust, model-agnostic generality across shifting generative distributions.

Existing approaches inherit the intrinsic limitations of pre-trained diffusion models, which primarily focus on salient objects, making it difficult to accurately edit small or less prominent regions. As shown in Tab.~\ref{tab:quantitative} and Fig.~\ref{fig:qualitative}, DiffEdit~\cite{couairon2022diffedit} and Blended-Diffusion~\cite{avrahami2023blended} produce rigid blends that fail to integrate naturally with the background, while MasaCtrl~\cite{cao_2023_masactrl} struggles to disentangle foreground and background layers. LEDITS++~\cite{brack2024ledits++} often introduces subtle artifacts in small-object edits, and CPAM~\cite{vo2025cpam} preserves background well but occasionally misaligns with the prompt. In contrast, FocusDiff achieves fine-grained and region-specific modifications while maintaining global structure and realism, delivering the best balance between text–image alignment (highest CLIPScore) and background preservation (lowest LPIPS), consistently outperforming all other state-of-the-art methods.

\begin{figure}[t!]
    \centering
    \includegraphics[width=\linewidth]{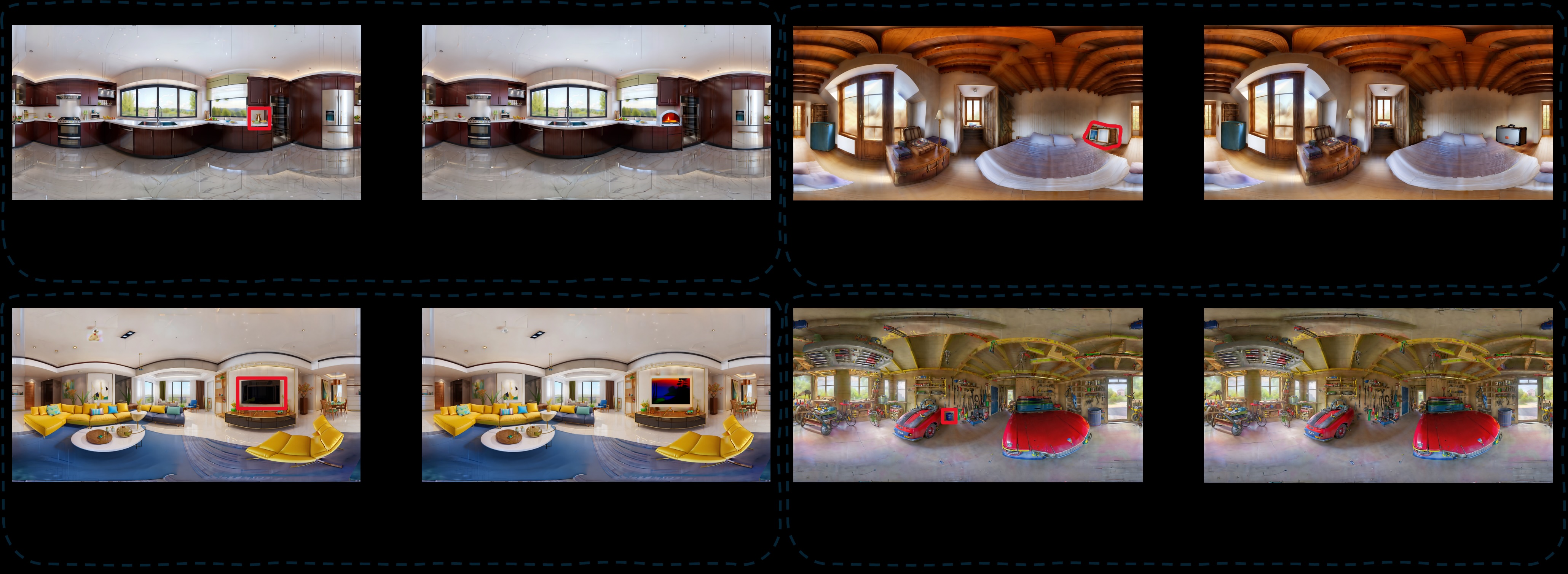}
    \vspace{-5mm}
    \caption{Qualitative evaluation of panoramic image editing using FocusDiff across four object replacement and four object removal cases.}
    \label{fig:placeholder}
    \vspace{-5mm}
\end{figure}

To evaluate the quality of panoramic edits produced by FocusDiff, we conducted eight test cases comprising four object replacement and four object removal scenarios, as shown in Fig.~\ref{fig:placeholder}. The results consistently align with the specifications provided in the input prompts. In the replacement tasks, masked objects are seamlessly substituted with the target entities described in the prompts while maintaining background integrity. In the removal tasks, objects are effectively eliminated using the proposed blurring mechanism. Furthermore, the region-based editing strategy confines modifications strictly to the target area, preserving global consistency and making FocusDiff particularly effective for panoramic image editing applications.

\subsection{Ablation study}

\begin{table}[!t]
\centering
\caption{Quantitative ablation results on the LIMB benchmark evaluating the performance impact of individual architectural components in FocusDiff.}
\label{tab:ablation_quantitative}
\begin{tabular}{lcc}
\toprule
\textbf{Configuration} & \textbf{CLIPScore} $\uparrow$ & \textbf{LPIPS} $\downarrow$ \\
\midrule
\textbf{FocusDiff-SD1.5 (Full Framework)} & \textbf{35.85} & \textbf{0.099} \\
FocusDiff w/o RCA (No Refocusing Cross-Attention) & 31.42 & 0.145 \\
FocusDiff w/o CPI (No Context-Preserving Integration) & 36.12 & 0.284 \\
Baseline (Direct Edit w/o Blurring Surrounding) & 29.80 & 0.312 \\
\bottomrule
\end{tabular}
\vspace{-5mm}
\end{table}

We conducted ablation studies to validate the effectiveness of Refocusing Cross-Attention (RCA), Context-Preserving Integration (CPI), and the importance of blurring the background, with quantitative metrics summarized in Tab.~\ref{tab:ablation_quantitative}. 
When both modules were enabled, the edited object was accurately localized and seamlessly blended into the scene, while the background remained consistent. Without RCA, the object’s identity and structural details were not transferred to the edited image, leading to a significant drop in CLIPScore to 31.42. Conversely, removing CPI severely degraded the background quality, causing the LPIPS error to surge to 0.284 due to a lack of spatial background features preservation during the diffusion process. Interestingly, while removing CPI slightly increased the CLIPScore to 36.12 because the editing area became unconstrained, it came at the cost of catastrophic background corruption and artifacts. Furthermore, directly editing without blurring the surrounding regions (Baseline) caused the model to frequently generate objects in the middle of the image rather than focusing on the intended region 
. Although those outputs were plausible, they failed to align with the desired position and yielded a poor balance of 29.80 CLIPScore and 0.312 LPIPS. These empirical findings demonstrate that both RCA and CPI are indispensable for balancing precise, text-aligned local editing with stable background consistency.

\subsection{User study}
\textbf{Participants. }
The user study involved 20 participants, consisting of 4 females (20\%) and 16 males (80\%). Participants’ prior experience with VR headsets varied:
\begin{itemize}
    \item 7 participants (35\%) reported no previous use.
    \item 7 participants (35\%) had limited exposure, using VR headsets rarely.
    \item 4 participants (20\%) indicated occasional use.
    \item 2 participants (10\%) regularly engaged with VR headsets.
\end{itemize}

This distribution of VR familiarity among participants helps support the reliability and broader applicability of the study’s findings.

\textbf{Setup and procedure. }
Our VR system was developed in Unity 6000.2.1f1 and executed on Meta Quest 2, with interactions supported by the Meta XR SDK (version 78). The server backend was deployed on an NVIDIA A100 server. In the user study, participants perform two tasks, including object removal and object replacement, on their selected image. After completing the tasks, they filled out the System Usability Scale (SUS) \cite{SUS}.

\begin{figure}[t!]
    \centering
    \includegraphics[width=0.8\linewidth]{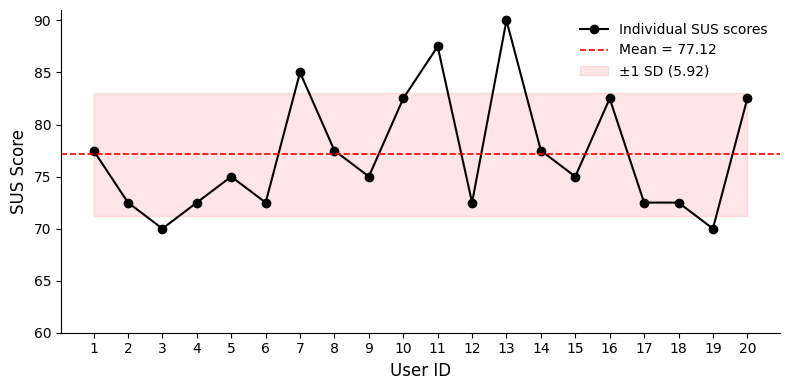}
    \vspace{-5mm}
    \caption{System Usability Scale (SUS) results show high usability for our VR-based panoramic editing system, with an overall score of 77.12/100.}
    \label{fig:userstudy_result}
    \vspace{-5mm}
\end{figure}

\begin{table}[t!]
\centering
\caption{Evaluation of our VR-based panoramic editing system using the System Usability Scale (SUS, $n=20$). Scales use a 1–5 rating from "1–strongly disagree" to "5–strongly agree".}
\label{tab:sus}
\begin{tabular}{p{0.7\textwidth}ccc}
\toprule
\textbf{Questions} & \textbf{Mean} & \textbf{SD} & \textbf{Skewness} \\
\midrule
1.  I think that I would like to use the VR system frequently. & 3.65 & 0.59 & 0.20 \\
2.  I found the VR system unnecessarily complex. & 1.75 & 0.72 & 0.39 \\
3.  I thought the VR system was easy to use. & 4.30 & 0.66 & -0.37 \\
4.  I think that I would need the support of a technical person to be able to use the VR system. & 2.85 & 0.93 & 0.70 \\
5.  I found the various functions in the VR system were well integrated. & 4.10 & 0.55 & 0.08 \\
6.  I thought there was too much inconsistency in the VR system. & 1.85 & 0.59 & 0.00 \\
7.  I would imagine that most people would learn to use the VR system very quickly. & 4.75 & 0.55 & -2.07 \\
8.  I found the VR system very cumbersome to use. & 1.45 & 0.60 & 0.93 \\
9.  I felt very confident using the VR system. & 4.20 & 0.62 & -0.11 \\
10. I needed to learn a lot of things before I could get going with the VR system. & 2.25 & 1.02 & 0.10 \\
\bottomrule
\end{tabular}
\vspace{-5mm}
\end{table}

\textbf{Results. }
The results, shown in Fig.~\ref{fig:userstudy_result} and Tab.~\ref{tab:sus}, indicate high usability of our system, with a SUS score of 77.12/100 (SD = 5.92). Among the positive criteria (odd-numbered items), four out of five received mean scores above 4.0, with the exception of the first item, which scored 3.65/5. For the negative criteria (even-numbered items), two items recorded mean scores above 2.0: 2.85 for item 4 and 2.25 for item 10.

This variation can be largely attributed to participants’ limited prior experience with VR. Approximately 70\% of users reported having never or only rarely used VR before. Consequently, during initial interactions, many participants required technical assistance or felt that additional guidance was necessary to use the system comfortably.


\section{Conclusion}

We presented FocusDiff, a tuning-free framework for precise region-based image editing. By combining refocused cross-attention with context preservation, it enables accurate local edits while maintaining background consistency. Experiments show that FocusDiff achieves high-fidelity results, handles small objects and complex scenes effectively, and outperforms existing methods. The framework also extends naturally to panoramic editing for VR and immersive media applications.

FocusDiff overcomes key limitations of pre-trained text-to-image diffusion models, enabling precise localized edits even for small objects. However, several limitations remain. The blur region is predefined and cannot be adjusted dynamically, limiting fine-grained control. Each edit also requires processing five image variants, increasing computational cost and inference time. In addition, the auto-connect algorithm struggles with incomplete or sparse user strokes. Since contour completion relies on simple linear interpolation, generated masks may not align with semantic boundaries, reducing interaction reliability in VR/AR settings. The framework also raises ethical concerns, including scene manipulation, privacy risks, and potential deepfake misuse. Future work will focus on improving efficiency, developing semantic-aware contour completion, and introducing safeguards for responsible deployment.


\textbf{Acknowledgments. }
This research is funded by Vietnam National University - Ho Chi Minh City (VNU-HCM) under Grant Number B2026-18-17.

\bibliographystyle{splncs04}
\bibliography{short}
\end{document}